\documentclass[letterpaper]{article}

\usepackage{natbib,alifeconf}  %% The order is important

\newcommand{\be}{\begin{eqnarray}}
\newcommand{\ee}{\end{eqnarray}}

\title{The structure of evolved representations across different substrates for artificial intelligence}
\author{Arend Hintze$^{1,2,3}$, Douglas Kirkpatrick$^{2,3}$ \and Christoph Adami$^{3,4,5}$ \\
$^1$Department of Integrative Biology, Michigan State University \\
$^2$Department of Computer Science and Engineering, Michigan State University \\
$^3$BEACON Center for the Study of Evolution in Action, Michigan State University\\
$^4$Department of Microbiology and Molecular Genetics, Michigan State University\\
$^5$Program in Ecology, Evolution, and Behavior Biology, Michigan State University\\
Corresponding author: hintze@msu.edu} % email of corresponding author

\begin{document}
\maketitle

\begin{abstract}
% Abstract length should not exceed 250 words
Artificial neural networks (ANNs), while exceptionally useful for classification, are vulnerable to misdirection. Small amounts of noise can significantly affect their ability to correctly complete a task. Instead of generalizing concepts, ANNs seem to focus on surface statistical regularities in a given task. Here we compare how recurrent artificial neural networks, long short-term memory units, and Markov Brains sense and remember their environments. We show that information in Markov Brains is localized and sparsely distributed, while the other neural network substrates ``smear'' information about the environment across all nodes, which makes them vulnerable to noise.
\end{abstract}

\section{Introduction}
The quest to recreate human-level intelligence within a computational substrate has gained traction in recent years, mostly due to the advent of deep learning methods and convolutional networks~\citep{schmidhuber2015deep,Bengioetal2015,Goodfellowetal2016}. While the success of these methods across a variety of different machine learning domains cannot be denied, their suitability as a method to implement artificial general intelligence has been questioned. In the visual categorization field, convolutional neural networks (CNNs) were found to display intriguing vulnerabilities~\citep{Szegedyetal2014}. In particular, the work of Szegedy et al.\ (and many subsequent papers, see for example~\cite{Nguyenetal2015}) found that it was possible to cause the network to misclassify an image by applying a certain imperceptible perturbation to the image, implying that the classification ability was not robust. Indeed, Szegedy et al. (2014) suggested that the vulnerability might be connected to the way that the semantic information about the objects to be discriminated resides across the entire neural space rather than in individual neural units. Jo and Bengio (2018)~\nocite{JoBengio2018} venture even further: they suggest that the extreme sensitivity of high performance CNNs to adversarial examples casts serious doubt that these networks are learning any high level abstractions in the dataset. In other words, when Jo and Bengio are observing that CNNs are sensitive to ``surface statistical regularities'' they are suggesting that CNNs are not intelligent at all; that they have no fundamental concept of the things that they are classifying. In the present work, we endeavor to shed some light on what it means for a network to have an understanding of the situation it is presented with, as opposed to a superficial reaction to the data set's regularities. In order to do this, we have to delve into the concept of {\em representations}.

\noindent {\bf Representations.}
In Machine Learning, the term ``representation'' refers to the internal encoding of data in terms of a feature vector. As such, the term ``representation'' describes a {\em transformation} of the image data (for example) into a form that is more suitable for classification. Generally speaking, these representations are meant to reduce the dimensionality of the underlying data space. One of the defining features of deep learning methods is that the optimal representations are not designed, but rather are automatically discovered, in a hierarchical manner. The term ``representations'' is, however, also used in a very different way within the field of cognitive science. In that field, the word ``representation" refers to {\em internal models} not of data, but of concepts in the world that are stored within the brain (see, for example,~\cite{johnson-laird,pinker}). Here (as in earlier work,~\cite{marstaller2013evolution}), we understand ``representations" to refer specifically to information about relevant features of the world encoded in the internal states of an organism or brain, information that goes {\em beyond} what is perceived in the agent's sensors (\cite{Haugeland1991,Clark1997a}). This implies in particular that representations can sometimes {\em misrepresent}~\citep{Haugeland1991}, quite unlike information present in sensors that always truthfully reflects the environment. Thus, in cognitive science, representations are context-dependent and refer to objects and concepts in the real world, as opposed to the ML term that instead refers to compressed versions of input data. In the following, we exclusively use the term ``representation" with the meaning from cognitive science, and explore how these internal models are stored quite differently within the artificial neurons or nodes of different computational substrates.

\begin{figure}[htbp] %  figure placement: here, top, bottom, or page
   \centering
   \includegraphics[width=2in]{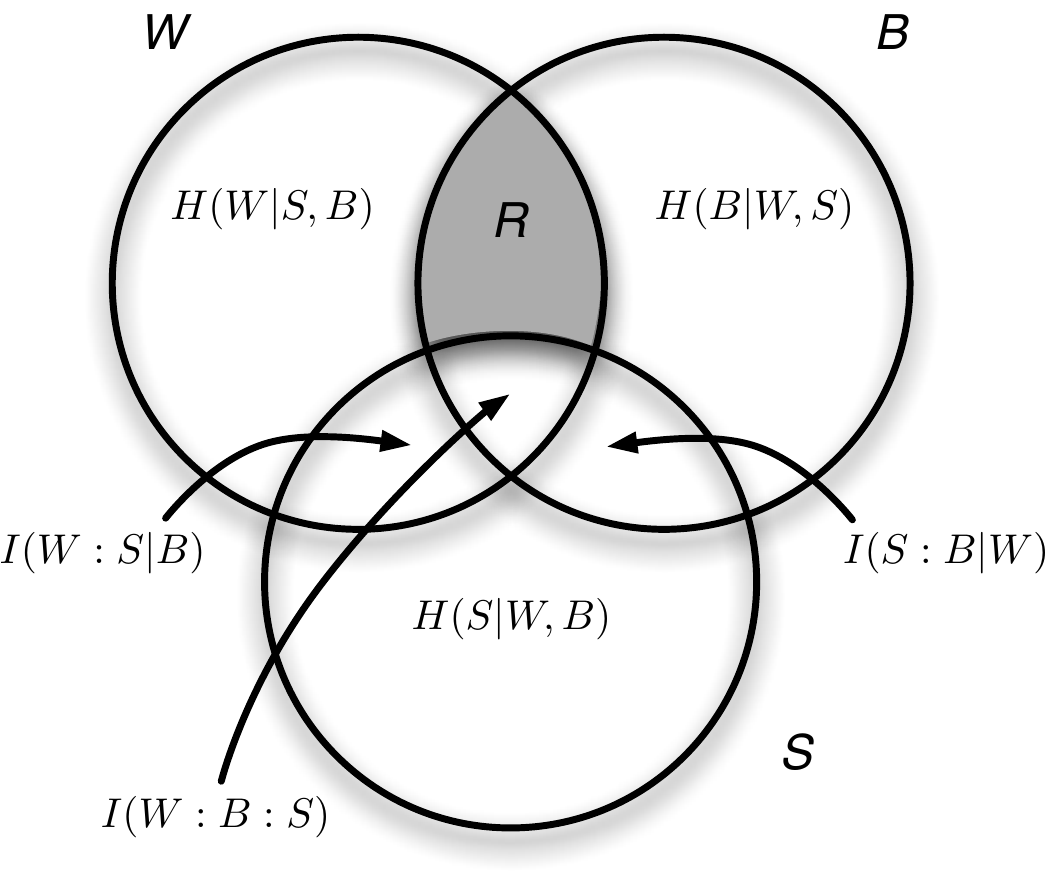} 
   \caption{Venn diagram of entropies and informations for the three random variables $W$, $S$, and $B$, describing the world, sensor, and agent internal (brain) states. The representation $R = H(W\! :\! B|S)$ is shaded.}
  % \caption{Venn diagram of entropies and informations for the three random variables $W$, $S$, and $B$, describing the world and sensor states, and agent internal degrees of freedom (i.e., brain states). The representation $R = H(W\! :\! B|S)$ is shaded.}
   \label{fig:Venn}
\end{figure}

\noindent{\bf Information theory.} To quantify the structure of representations, we need to be able to measure them. In previous work~\citep{marstaller2013evolution} we succeeded in giving an information-theoretic foundation to the term representation, as the information that internal brain states have about concepts in the world {\em given} the sensory data. By inspection of the information-theoretic Venn diagram relating the entropies of world states $W$, internal brains states $B$, and sensor states $S$ (see Fig.~\ref{fig:Venn}), the representation $R$ (how much the brain knows about the world given the sensory data) can be written as 
\be
R=H(W:B|S)=H(W:B)-I(W:B:S)\;. \label{eqR}
\ee
In Eq.~(\ref{eqR}), $H(W\!:\!B)$ refers to the shared Shannon entropy between world states and brain states, and $I(W\!:\!B\!:\!S)$ stands for the information shared between world, brain, and sensors, something that Phillips and Singer (1997)\nocite{PhillipsSinger1997} have called ``coherent information"\footnote{See Phillips and Singer (1997) or Marstaller et al. (2013) for more details on how to calculate the relevant information-theoretic terms going into the definitions used here.}.
But while Phillips and Singer assume that evolutionary processes have maximized coherent information~\citep{Phillipsetal1994}, it turns out that this quantity is for the most part {\em negative}, and it is instead $R$ that is maximized~\citep{marstaller2013evolution}.
%Fig. 1

\noindent{\bf Brains, Dynamical Worlds, and Evolution.} While a significant fraction of work in the Deep Learning field deals with the classification of static scenes (excepting Mnih et al., 2015\nocite{Mnihetal2015}, and similar work on Deep Q-Learning) vertebrate brains must compute sensory outputs in time, in a constantly changing world. Furthermore, in such dynamic worlds there are no ``correct actions" that a supervised training algorithm can use. While Deep Q-Learning algorithms can automatically generate value functions that are near-optimal and change over time, they have difficulty with long-range planning, and are likely to be as vulnerable to adversarial attacks as their static CNN counterparts. Here, we take a different approach. We focus on a behavioral task in a dynamic world that requires memory (the Active Categorical Perception Task, described below), and use Darwinian evolution to optimize the control structures. We will focus on three different computational substrates that are capable of memory: recurrent neural networks (RNNs), Long Short-Term Memory (LSTM) networks~\citep{schmidhuber2015deep}, and Markov Brains~\citep{hintze2017markov}, and compare the structure and distribution of their internal representations. We note that the Hierarchical Temporal Memory (HTM) approach to cognitive computing~\citep{HawkinsBlakeslee2004,GeorgeHawkins2009} shares many properties with Markov Brains (by design), but we do not investigate their properties here.

\section{Material and Methods}
\noindent {\bf Active Categorical Perception Task.}
In this classic task~\citep{Beer1996,Beer2003,vandarteldiss,vandarteletal05,marstaller2013evolution,Albantakisetal2014} a mobile agent has to catch or avoid blocks that move towards it. The agent sits on a rail and can only move left or right, with periodic boundary conditions. The stage is 16 steps (units) wide and blocks are dropped from the top, which is 32 steps away. At every update, the blocks move one step closer but also sideways to the left or right. Blocks of size two have to be caught, while blocks of size four have to be avoided.
%Fig. 2
\begin{figure}[hbt]
\begin{center}
  \includegraphics[width=0.75\columnwidth]{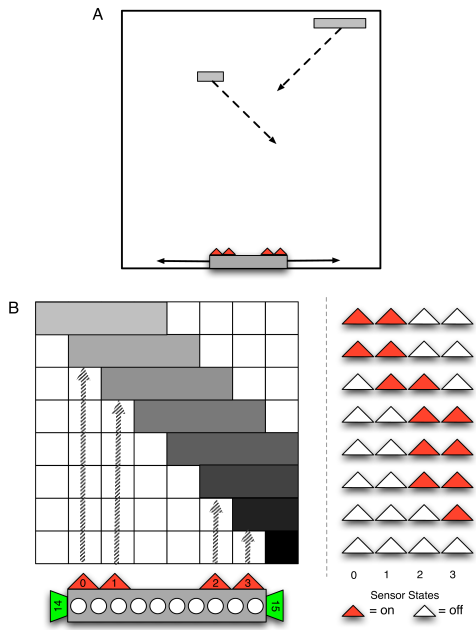} 
\end{center}
\caption{A: Large or small blocks fall one at a time diagonally toward the bottom row of a $16\times32$ discrete world, with the agent on the bottom row. In this illustration, a large brick is falling to the left, while a small brick is falling to the right (in simulations, only one block is falling at the time, and any one brick can fall to the left or to the right). In this world, agents are rewarded for catching small blocks and punished for catching large blocks. B: A depiction of the agent's states (bottom left: triangles depict sensors, circles illustrate brain (internal) states, trapezoids denote actuators) and the sequence of activity patterns on the agent's 4-bit retina (right), as a large brick falls to the right. Reproduced from~\cite{marstaller2013evolution}, with permission.
}
\label{fig:TaskIllustration}
\end{figure}
The agent perceives the environment and approaching blocks by using four upward facing sensors. The agent is six units wide, and has two sensors on the left and two on the right side of its body leaving a blind spot in the middle. In order to perceive the size and direction of the approaching block properly, the agent has to first maneuver to an appropriate location, observe the block for a couple of updates, and then make a decision whether to catch or avoid the block (see Figure \ref{fig:TaskIllustration}).
When evolving an agent to perform this task, we assess performance using all 64 possible start conditions (2 sizes, 2 directions, 16 start locations). For scoring, we use an exponential fitness function that multiplies the score by $1.05$ for every successful action, and divides the score by $1.05$ for every mistaken action (see Marstaller et al., 2013, for more details). 

\noindent{\bf Markov Brains.}
Markov Brains are networks of logical elements that connect inputs and outputs via internal states~\citep{hintze2017markov}. Traditionally, the logical elements are deterministic or probabilistic logic gates. Here we only use deterministic logic gates, which we set up to have between 1 and 4 inputs and between 1 and 4 outputs. These gates read from and write into inputs, outputs, and hidden states. Note that writing into an input has no effect.

A genome is used to encode the logic and connectivity of each gate. Point mutations (implemented with a per-site mutation rate of $0.005$), deletions ($p=0.0002$ times genome length), and gene duplications ($p=0.0002$ times genome length) are applied every time an offspring is created by the genetic algorithm to populate the next generation. To begin evolution, genomes containing 5,000 sites are generated randomly (genome size is constrained to between 5,000 and 20,000 sites). Duplications and deletions occur in chunks of 256 to 512 sites, with the number of sites drawn from a uniform random distribution. The task requires four sensor inputs, two motor outputs, and we use 10 hidden states. For a more  detailed description of Markov Brain technology, see~\citeauthor{hintze2017markov} (\citeyear{hintze2017markov}). All computational evolutionary experiments were performed using the MABE framework~\citep{bohm2017mabe}. 

\begin{figure}
\begin{center}
  \includegraphics[width=0.85\columnwidth]{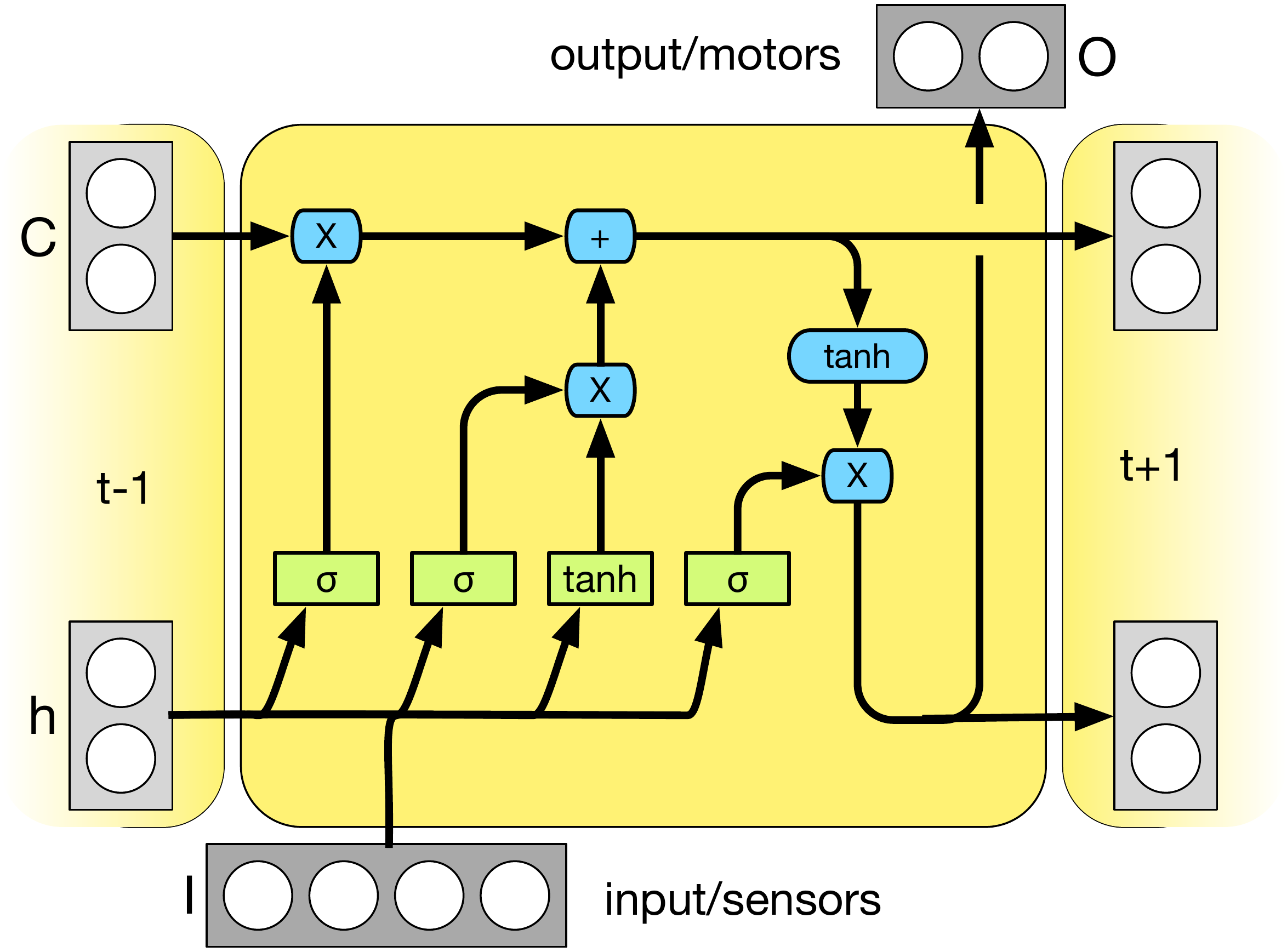} 
\end{center}
\caption{ Overview of an LSTM. The input nodes $I$ are joint with a set of recurrent states $h$. These joint states are used in four different computations. Like in a regular ANNs these inputs are multiplied with four different sets of weights and aggregated using a summation function. The results are then used in four different activation functions (green boxes, sigmoid $\sigma$, hyperbolic tangent $\tanh$). The resulting output vectors are further multiplied (blue box X) or added (blue box +) with each other or another set of recurring hidden states $C$. The outputs $O$ and the two new recurring hidden states $C_{t+1}$ and $h_{t+1}$ are now the result of the prior computations.
}
\label{fig:LSTMoverview}
\end{figure}

\noindent{\bf LSTM networks.}
Long-short-term-memory~\citep{HochreiterSchmidhuber1997} artificial neural networks (LSTM) implement recurrence in a special way (see Figure \ref{fig:LSTMoverview}). Instead of ``just" looping outputs back to inputs, the LSTM uses two different streams of recurrence ($h$ and $C$) as well as more complex modification and internal update rules, and as such greatly deviate from a classical layered model of ANNs~\citep{russell2003artificial}. LSTMs have been shown to solve tasks that require memory (recurrence) very well when using Deep Learning~\citep{schmidhuber2015deep}. Here, like in the MBs, we use a genome to encode each weight required for the LSTM. The genome for the LSTM networks can mutate in the same ways as for the MBs. We use the same number of inputs and outputs as for the MBs, but have the number of hidden states ($C$ and $h$) set to ten in order to have a comparable hidden state space. Note that LSTMs use continuous and not binary values. 
%Fig. 3

\noindent{\bf Recurrent ANNs.}
An artificial neural network is typically organized in layers of nodes, where the top layer receives inputs and the bottom layer is interpreted as outputs \citep{russell2003artificial}. In between can be arbitrarily many and arbitrarily large hidden layers, but all nodes from one layer are always connected to all nodes of the next layer. The state of each node after the input layer is defined by a transfer function and a threshold function. Here we use a simple perceptron rule, where the transfer function for each node is the sum of all inputs times all weights, and the threshold function is the hyperbolic tangent.

\begin{figure}
\begin{center}
  \includegraphics[width=0.85\columnwidth]{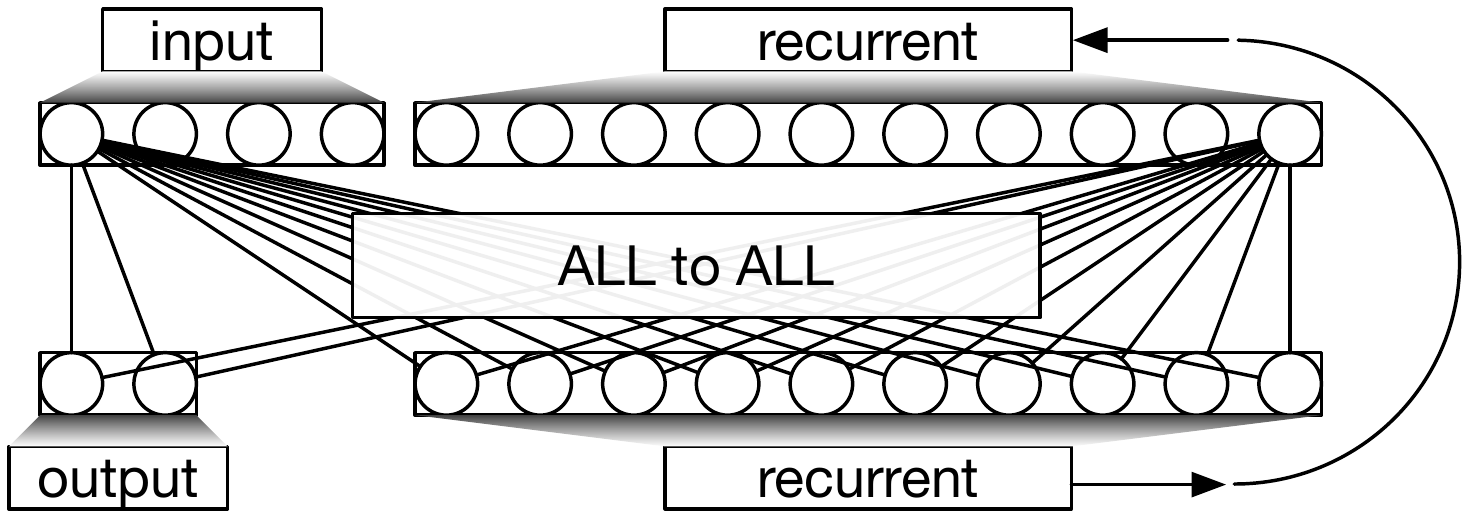} 
\end{center}
\caption{Overview of the RNN used. Four input nodes are joined with 10 recurring hidden nodes. The next layer is computed by summing over all inputs and weights and applying a hyperbolic tangent function for all nodes of the next layer. The first two nodes of the next layer serve as the outputs, while the remaining 10 nodes are used to implement the recurrence.
}
\label{fig:ANN}
\end{figure}

In order to make this ANN recurrent (i.e., an RNN), $10$ extra nodes are added to the input and output layer. After each update of the ANN the content of the recurrent nodes is copied from the output layer back to the input layer (see Figure \ref{fig:ANN}).
%Fig. 4

\noindent{\bf Local and global representations.}
In order to quantify $R$ we have to record the states of the sensors, brain (internal) states, as well as the state of the environment over time. While we can simply record the sensor and hidden (brain) states at every update, the world states require more attention. Although the world as a whole is in a specific state at every time point, there is no obvious way to meaningfully capture that the entirety of that state. We instead coarse-grain the world states into significant concepts thought to be relevant to the agent. From prior experiments~\citep{marstaller2013evolution,schossau2015information} we know that in order for the agent to solve the task it needs to know whether the block is small or large, if the block's position is to the agent's left or to the right, and whether the block is moving to the left or right. This allows us to use these three categories to define digital random variables $W_s$, $W_l$, and $W_d$, encoding block size, location, and direction, respectively, and use them to define a coarse-grained world variable.  Once all those states have been recorded, we can consider sensors $S$, brain states $B$, and world states $W$ as independent random variables.  This definition allows us to compute the total amount of information about the world stored in brain states given the information from the sensors, which is $R$. 

To determine how this knowledge (the knowledge about the world stored in the representations) is distributed across nodes, we also partition the brain states into the product random variable $B=\prod_{i=1}^{10} B_i$, one for each hidden or recurrent node in the brain. This allows us to define concept-specific representations in particular nodes, for example how much of the concept of block size is represented in node $i$, as the representation matrix $M_{{\rm s},i}= H(W_s\!:\!B_i|S)$. By mapping out the representation matrix in concept x node space, we can capture whether representations are smeared out over all nodes (global), or whether they are localized to particular brain regions. 

\begin{figure}
\begin{center}
  \includegraphics[width=0.55\columnwidth]{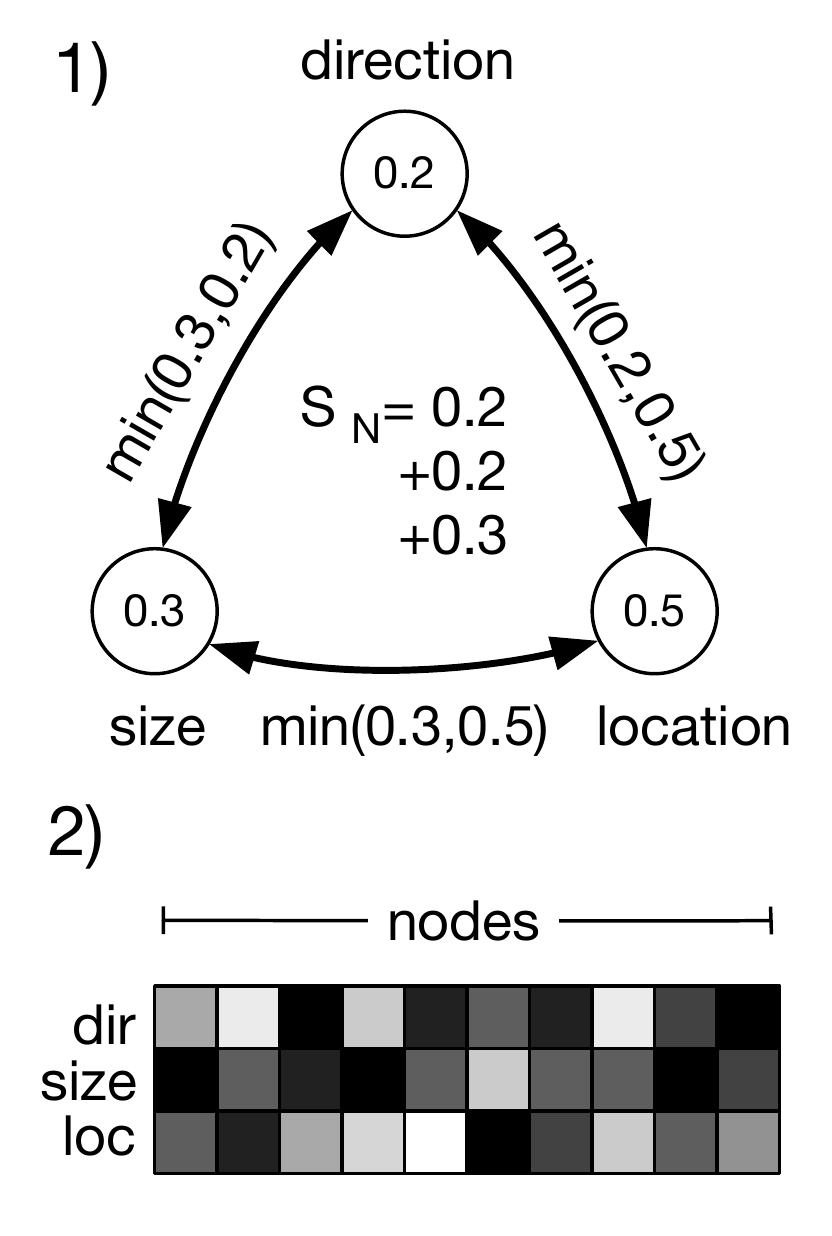} 
\end{center}
\caption{Illustration of how smearedness is computed. 1) A node stores different amounts of representations about the three concepts (direction $0.2$ bits, location $0.5$ bits, size $0.3$ bits). The arrows indicate all pairwise comparisons, and for each comparison the minimum of both values defines the overlap in representations. All resulting values are summed. 2) Representation matrix. Each row of the matrix $M$ is a different concept, each column a different node.
}
\label{fig:smearOverview}
\end{figure}

\noindent {\bf Smeared Representations.}
The representation Matrix shows how much information each node has about each concept. A hand-designed brain that could solve the block catching task would probably take advantage of discrete mappings between nodes and concepts. Specifically, such a brain would probably have one node per concept, while the other nodes would be used to perform other computations necessary to solve the task. The representation matrix $M$ would consequently have very sparse and discrete representations, while the rest of the matrix would be empty. As we will show, evolved computational systems have smeared representations, meaning that nodes store information about multiple concepts at the same time (``concept-smearedness" $S_{C}$), and representations about concepts are also smeared over multiple nodes (``node-smearedness"  $S_{N}$). As these measures of smearedness are not universally we defined, we offer a definition below. If a node has only representations about one concept and no other, the information is discrete and not smeared. Consider, in contrast, a node that has $0.2$ bits of information about the direction, $0.3$ bits of information about the size, and $0.5$ bits of information about the location of the block. This node by our definition carries a smeared concept representation, and these representations may also be smeared across multiple nodes. The overlap between each concept pair is the minimum of the two values, and the total amount of smearedness is thus the sum over all pairwise minima (see Figure \ref{fig:smearOverview}).
%Fig. 5

This allows us to quantify how smeared representations are across nodes (columns), by summing over all nodes $i$ and for all combinations of concepts $j$ and $k$:

\begin{equation}
    S_{N}=\sum_{i}\sum_{j>k}min(M_{ji},M_{ki})
    \label{equ:nodeSmear}
\end{equation}

Similarly, we compute the smearedness of concepts across nodes as the sum over all concepts $i$ for all combination of nodes $j$ and $k$:
\begin{equation}
    S_{C}=\sum_{i}\sum_{j>k}min(M_{ij},M_{ik})
    \label{equ:conceptsSmear}
\end{equation}

\noindent {\bf Robustness.} Later, when considering the quality and dispersion of representations, we must also assess how robust these brains and their representations are to external noise. To measure how robust each brain is against noise, each evolved brain is tested over a range of possible noise levels applied to the inputs. At each update, each sensor has a probability $p$ to receive a random input of $0$ or $1$ instead of the input it would otherwise receive from the environment. Each brain is tested 20 times for each different degree of noise, and the performance is averaged over all replicates. A perfectly robust brain should be able to tolerate a high degree of noise before performance drops, while fragile brains will lose performance at the smallest level of noise.

\section{Results.}
\begin{figure}
\begin{center}
  \includegraphics[width=0.85\columnwidth]{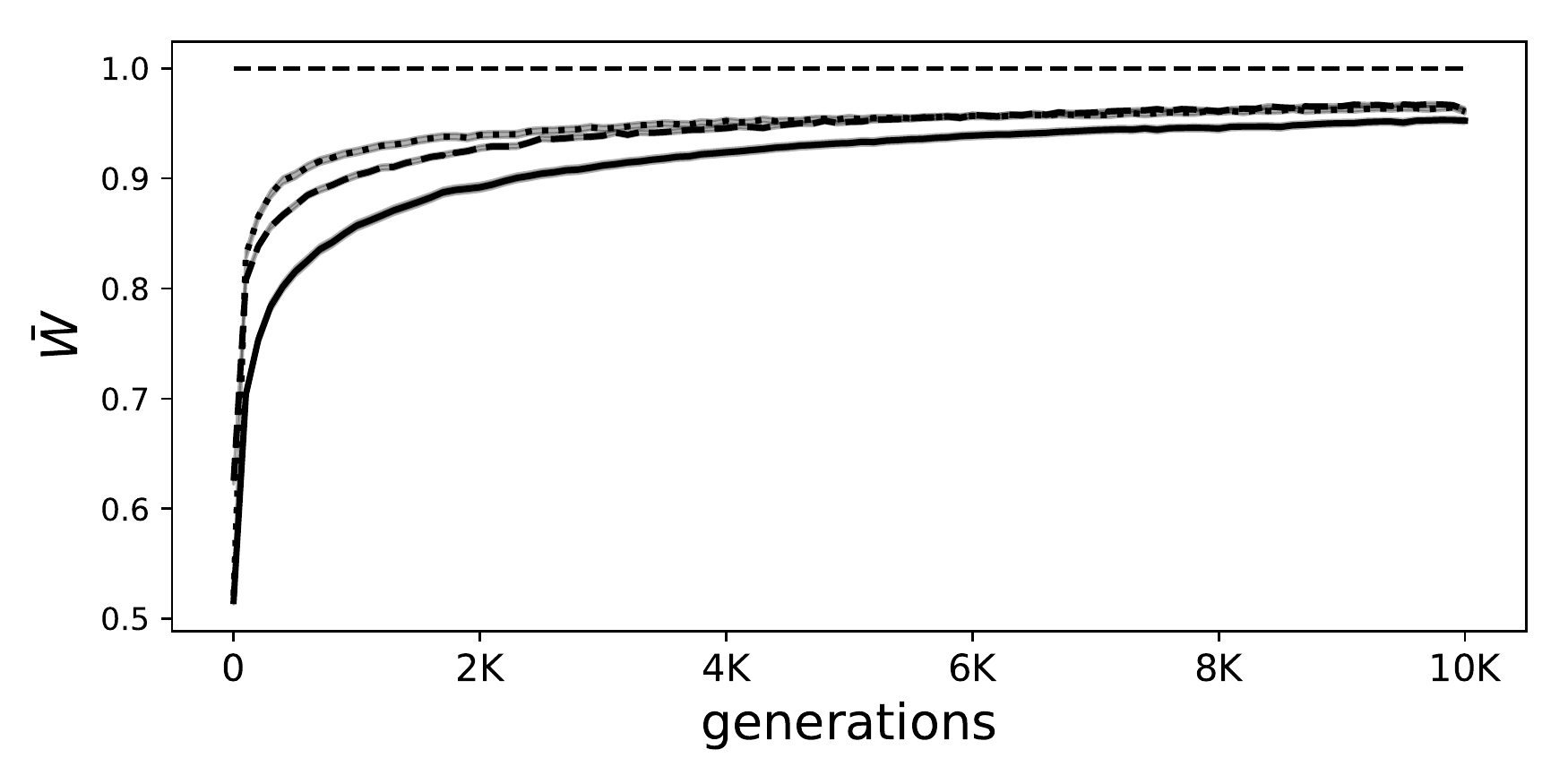} 
\end{center}
\caption{Evolution of performance. The mean fitness $\bar{W}$ of Markov Brains (solid line), LSTMs (dashed line), and RNNs (dotted line) over the line of descent (generations). The horizontal dashed line indicates optimal performance (64 correct choices out of 64). The gray shadows represent the standard deviation. Averages are generated over all 400 replicate experiments per brain type.}
\label{fig:fitness}
\end{figure}

\begin{figure}
\begin{center}
  \includegraphics[width=0.85\columnwidth]{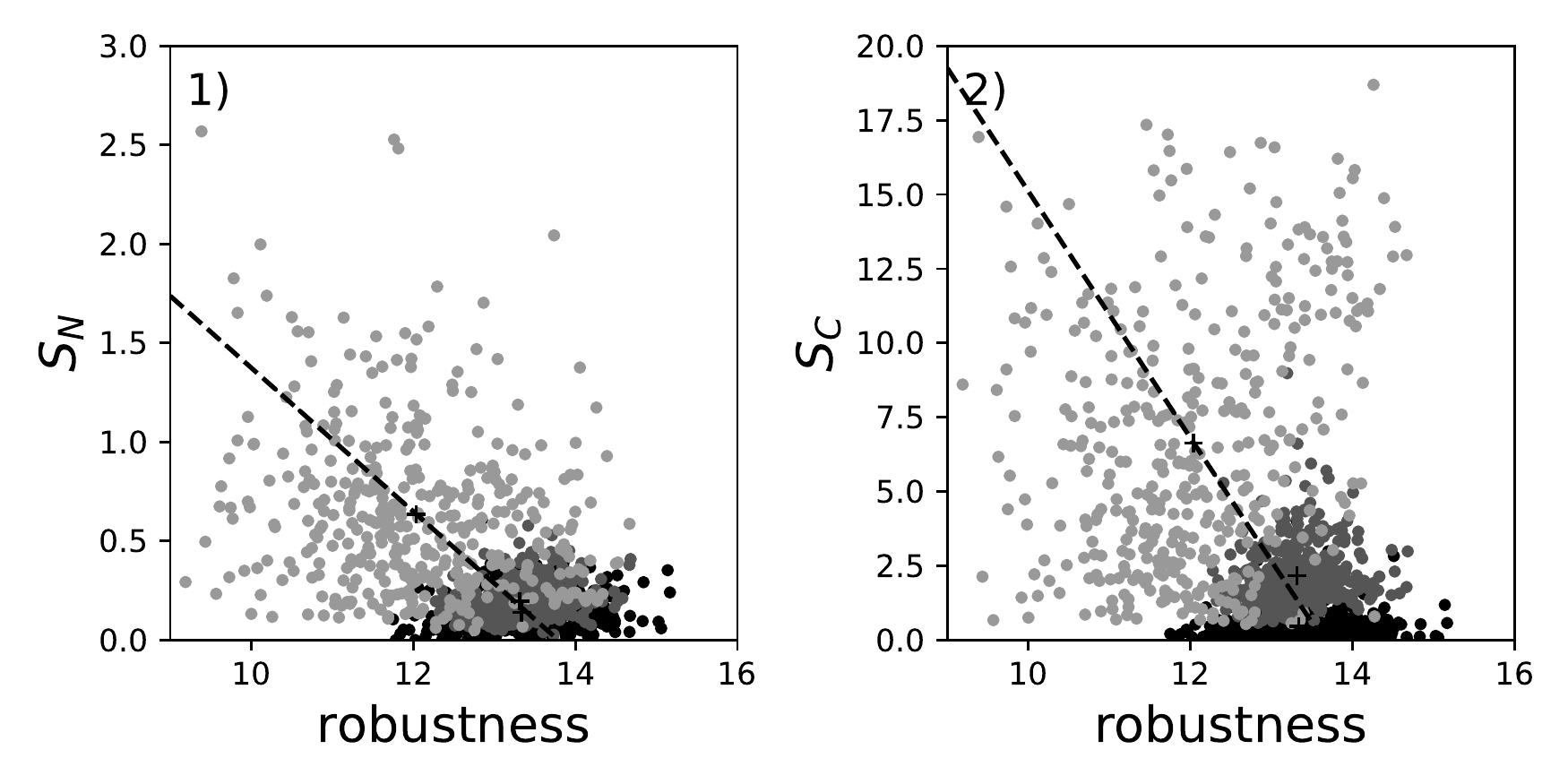} 
\end{center}
\caption{Robustness versus smeardness of representations. The smearedness of representation (y-axis) for nodes ($S_N$, panel 1) and concepts ($S_C$, panel 2) for all evolved types of brains: Markov Brain (black dots), LSTM (dark gray), RNN (light gray) against robustness ($x$-axis). The '+' indicate the mean smeardness over mean robustness for each of the three brain types. The dashed line is a linear regression fit for those means, indicating that the more robust brains are, the less smeared their representations are.}
\label{fig:smeardnessVsRobustness}
\end{figure}

We evolved 400 independent populations of 100 agents for each of the three brain types (MB, LSTM, and RNN) for 10,000 generations. After that, the line of descent~\citep{lenski2003evolutionary} was reconstructed, and we confirm that evolution converged appropriately (see Figure~\ref{fig:fitness}) over the first 7500 generations, meaning that further increase in performance is minimal thereafter. Of those 400 replicate runs, 154 MBs, 64 of the LSTMs, and 130 of the RNNs attained perfect performance. For the rest of this analysis, we will only use those optimally performing agents (with the exception of Figure~\ref{fig:smeardnessVsRobustness}).
%Fig. 6

\begin{figure}
\begin{center}
  \includegraphics[width=0.85\columnwidth]{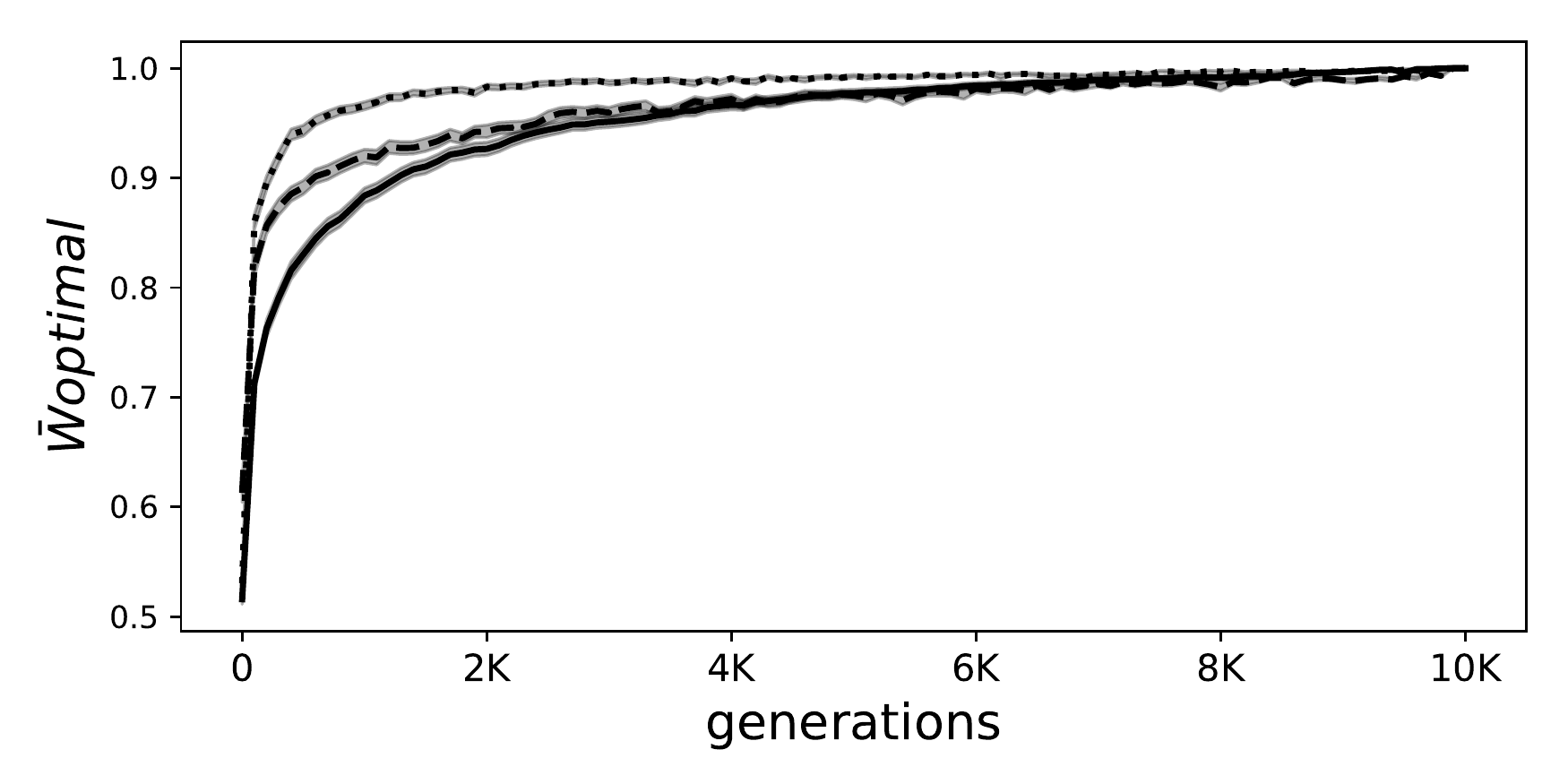} 
\end{center}
\caption{Evolution of fitness for those individuals that evolve to be optimal performers. The mean fitness $\bar{W}$ of Markov Brains (solid line), LSTMs (dashed line), and RNNs (dotted line) over the line of descent (generations). Averages are generated over those replicates that resulted in an optimal performer.}
\label{fig:optimalFitness}
\end{figure}

\begin{figure}
\begin{center}
  \includegraphics[width=0.85\columnwidth]{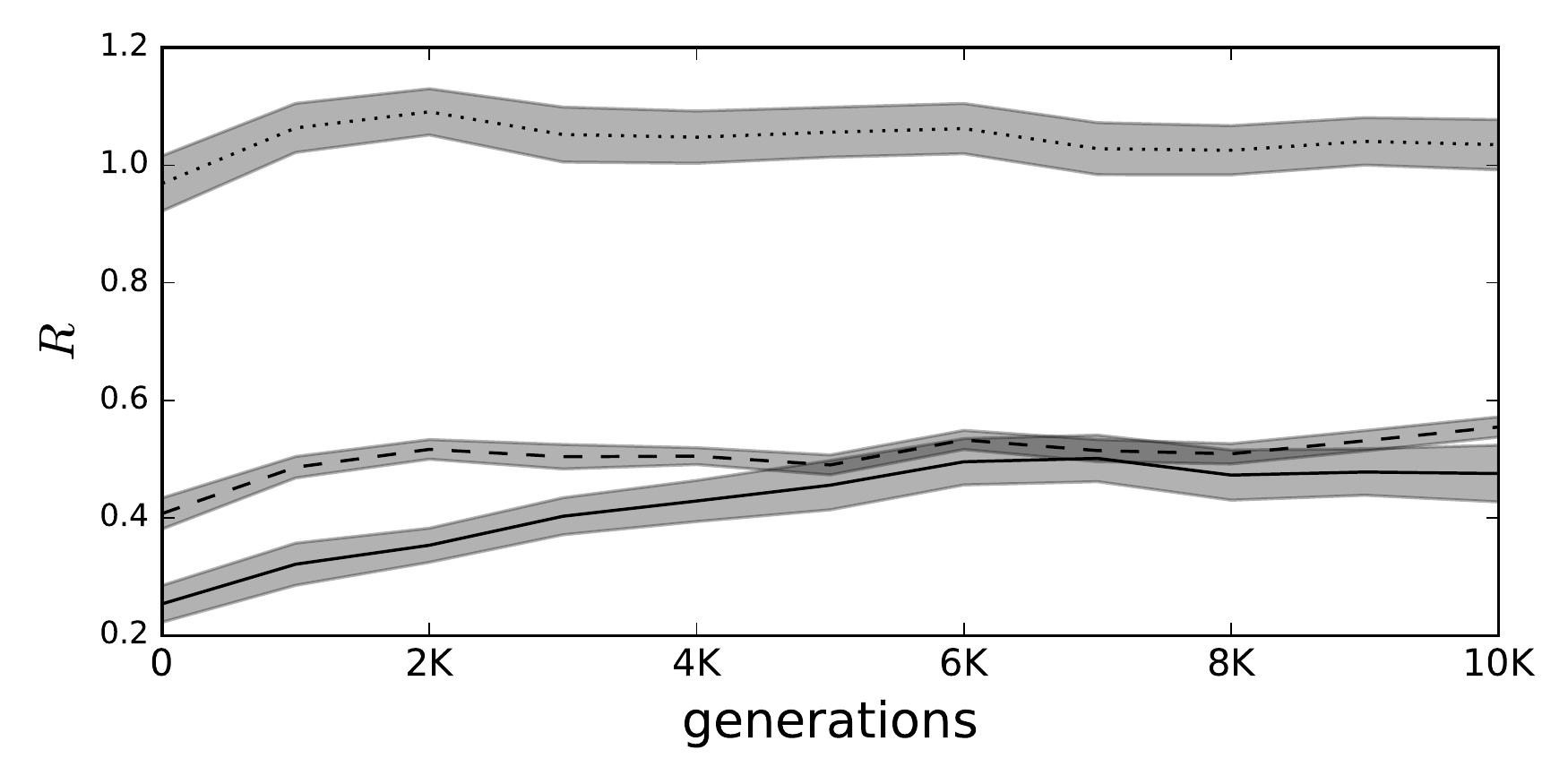} 
\end{center}
\caption{Evolution of representations. The mean $R$ of 30 randomly chosen individuals that evolve to become optimal performers, over generations: Markov Brains (solid line), LSTMs (dashed line), RNN (dotted line). Standard error is indicated as a grey shadow.}
\label{fig:rOverTime}
\end{figure}

We find that over the course of evolution, LSTMs and RNNs initially adapt faster than MBs but seem to struggle slightly to achieve maximum performance (see Figure~\ref{fig:optimalFitness}). However, these differences might be explained by the different effects mutations to the genome have in the different systems.
%Fig. 7
We find that the total amount of representation for the top performers increases over evolutionary time as expected (see Figure \ref{fig:rOverTime}). We observed before~\citep{marstaller2013evolution} that RNNs change the amount of representation only slightly over time. The RNN is designed to maintain information about the environment, and thus probably has at least some form of internal information, but must evolve to use these representations. The LSTM and MB on the other hand must first be optimized to carry representations, and then evolve to integrate them into their future decisions.
%Fig. 8

\begin{figure}
\begin{center}
  \includegraphics[width=0.85\columnwidth]{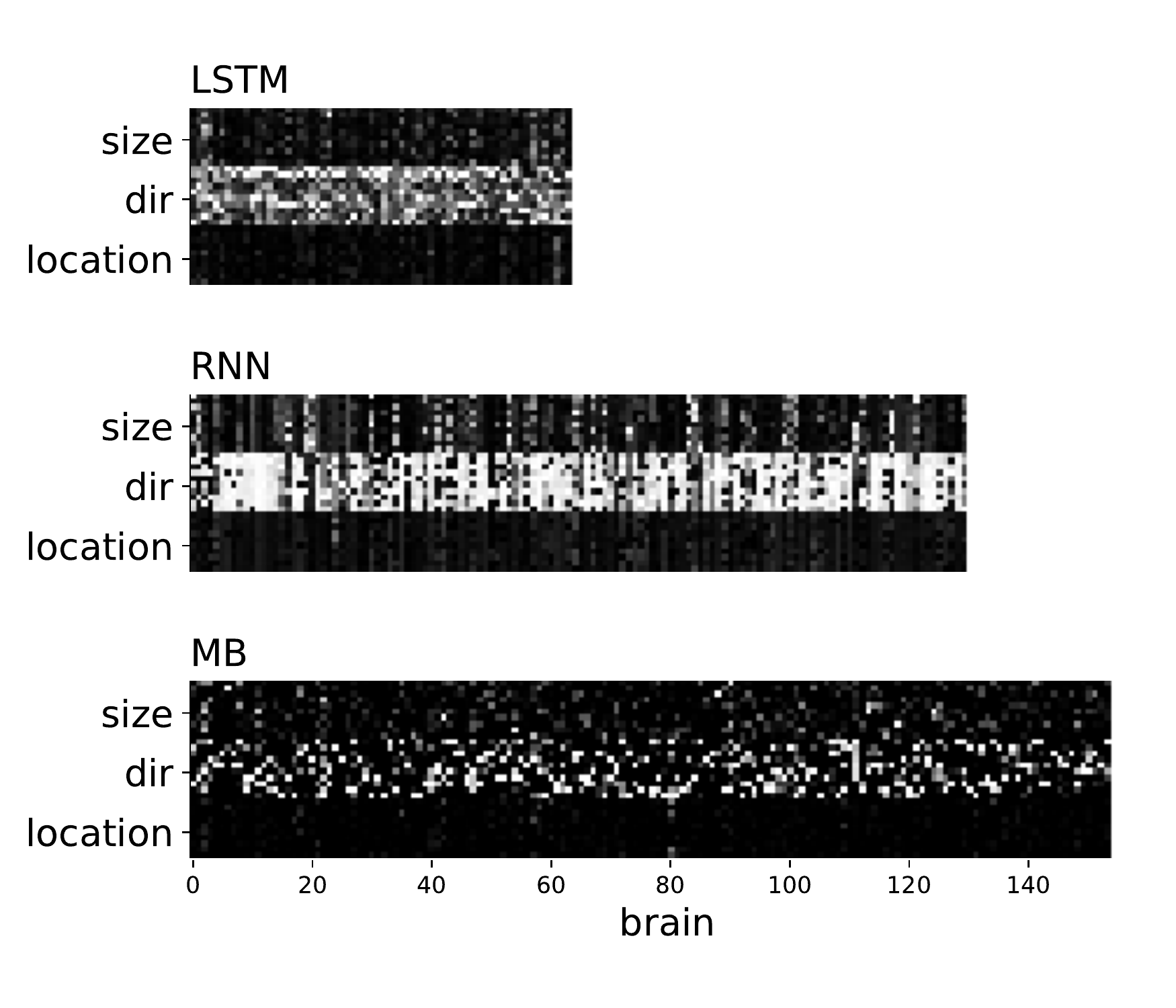} 
\end{center}
\caption{Comparison of representations across different brain types. For each optimally performing brain we rearrange the $M$ matrix into a vector, where the first 10 elements now contain the information of the 10 hidden states about the size (size) of the falling block, the next 10 are about the direction (dir) of the block falling, and the last 10 are about the information if the block is to the right or left (location). These vectors are stacked next to each other forming each of the panels of the figure, corresponding to their brain type. The normalized relative values of $R$ derived from the $M$ are coded in grey, the brighter the higher the value, black for $0.0$}
\label{fig:comparison}
\end{figure}

Independently of the amount of each representation each brain type has, the brain types might represent the environment differently. We therefore analyzed the amount of representation each node has about each concept individually. This creates a matrix ($M$) where each row reflects a concept, and each column a hidden node of the system. Thus, each element of this matrix contains the information a specific node has about a specific concept. By visual inspection, it seems as if the representations in the MB are much more distinct, while in the LSTMs and even more so in the RNNs representations are much more ``smeared'' or distributed (see Figure \ref{fig:comparison}). We have already seen that multiple nodes can be used to represent a concept together (sparsely distributed representations), however the extent to which LSTMs and RNNs distribute their representations is surprising to us. 

The difference in representation distribution might be explained by the way these brains work internally. In a MB the logical elements can connect arbitrarily, and the topology of the entire network can evolve, whereas in an LSTM and RNN the topology is predefined, and generally speaking, everything is connected to everything else. This connectivity is similar to a layer in a classic ANN where all nodes of one layer are connected to all nodes of a following layer.
%Fig. 9

\begin{figure}[htbp]
\begin{center}
  \includegraphics[width=0.85\columnwidth]{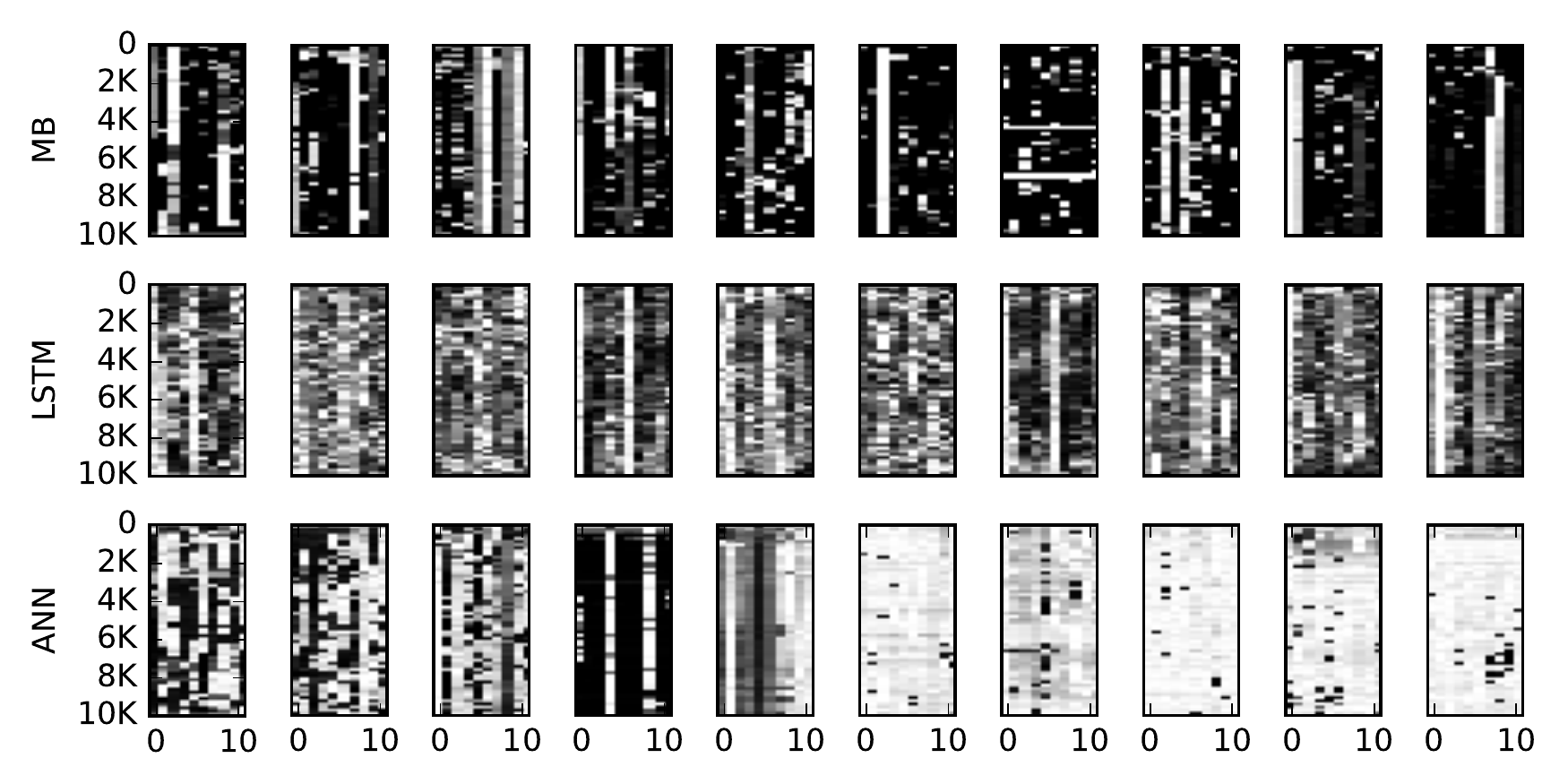} 
\end{center}
\caption{Evolution of representation. Each row represents a different type of brain, top row Markov Brains, middle row LSTM, and bottom row RNNs. Each panel shows the linearized representation matrix $M$ for all generations between $0$ and $10,000$ in increments of $100$. Here only 10 randomly chosen brains from each category are shown.}
\label{fig:rEvolution}
\end{figure}

We also find that how representations evolve is different between systems. When comparing, for example, how representations about the concept of the direction the blocks are falling changes over the course of generations, the difference becomes most apparent (see Figure \ref{fig:rEvolution}). In Markov Brains specific hidden states are chosen early in the evolutionary process to contain the information about the direction (see Figure \ref{fig:rEvolution} top row). In RNNs more or less all states are used to store information about the direction, even though occasionally a couple of states seem to be preferred (see Figure \ref{fig:rEvolution} bottom row). LSTMs fall between these two extremes (see Figure \ref{fig:rEvolution} middle row). They have a few states containing information about specific concepts in the environment, while still other states have additional information ``smeared''. In addition, LSTMs constantly change over the course of evolution how strongly their hidden states represent concepts. This is not surprising when one considers that mutations affect weights, which themselves change the behavior of a single node, but because in the LSTM many things are interconnected, these changes affect the entire network. In a MB on the other hand, changes to the logic of a gate are much more contained, affecting only those states that are directly affected by the connections via gates.
%Fig. 10
%Fig. 11
\begin{figure}[htbp]
\begin{center}
  \includegraphics[width=0.85\columnwidth]{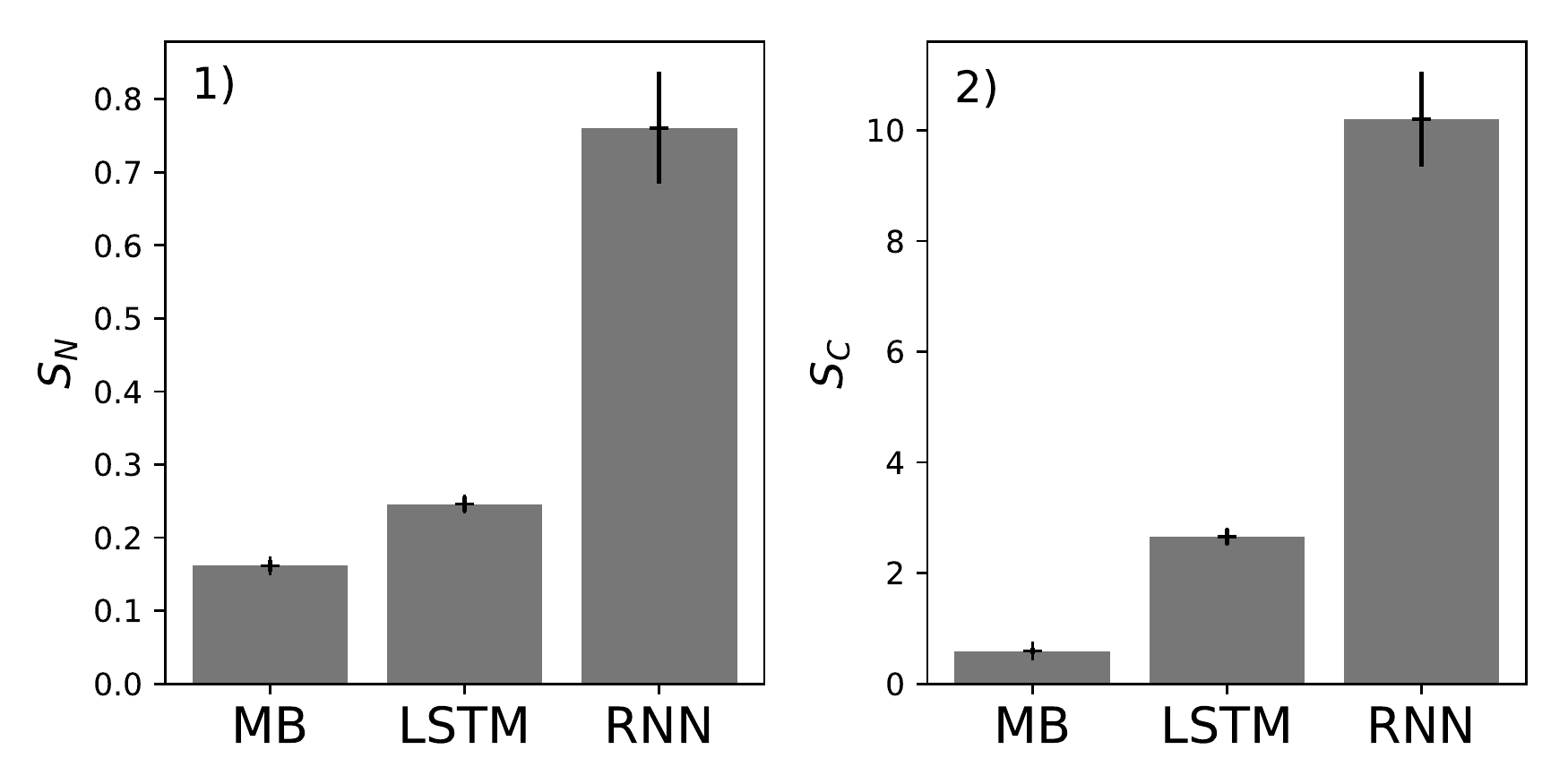} 
\end{center}
\caption{Smearedness of representations. Average smearedness of representations for Markov Brains (MB), LSTMs, and RNNs of optimal performers. Panel 1) shows how much representations are smeared over each node (Equation \ref{equ:nodeSmear}), Panel 2) shows how much representations are smeared over concepts (Equation \ref{equ:conceptsSmear}). Error bars indicate the standard error.
}
\label{fig:smeardnessComparison}
\end{figure}

To quantify how smeared representations are, we compute the overlap between concepts or nodes across the representation matrix $M$ for all optimally performing agents (Equation \ref{equ:nodeSmear} and \ref{equ:conceptsSmear}). We find that representations are smeared across concepts as well as nodes, however in Markov Brains we find the least smeared representations, followed by LSTMs, while in RNNs representations are smeared the most for both nodes and concepts (see Figure \ref{fig:smeardnessComparison}). 

\begin{figure}
\begin{center}
  \includegraphics[width=0.85\columnwidth]{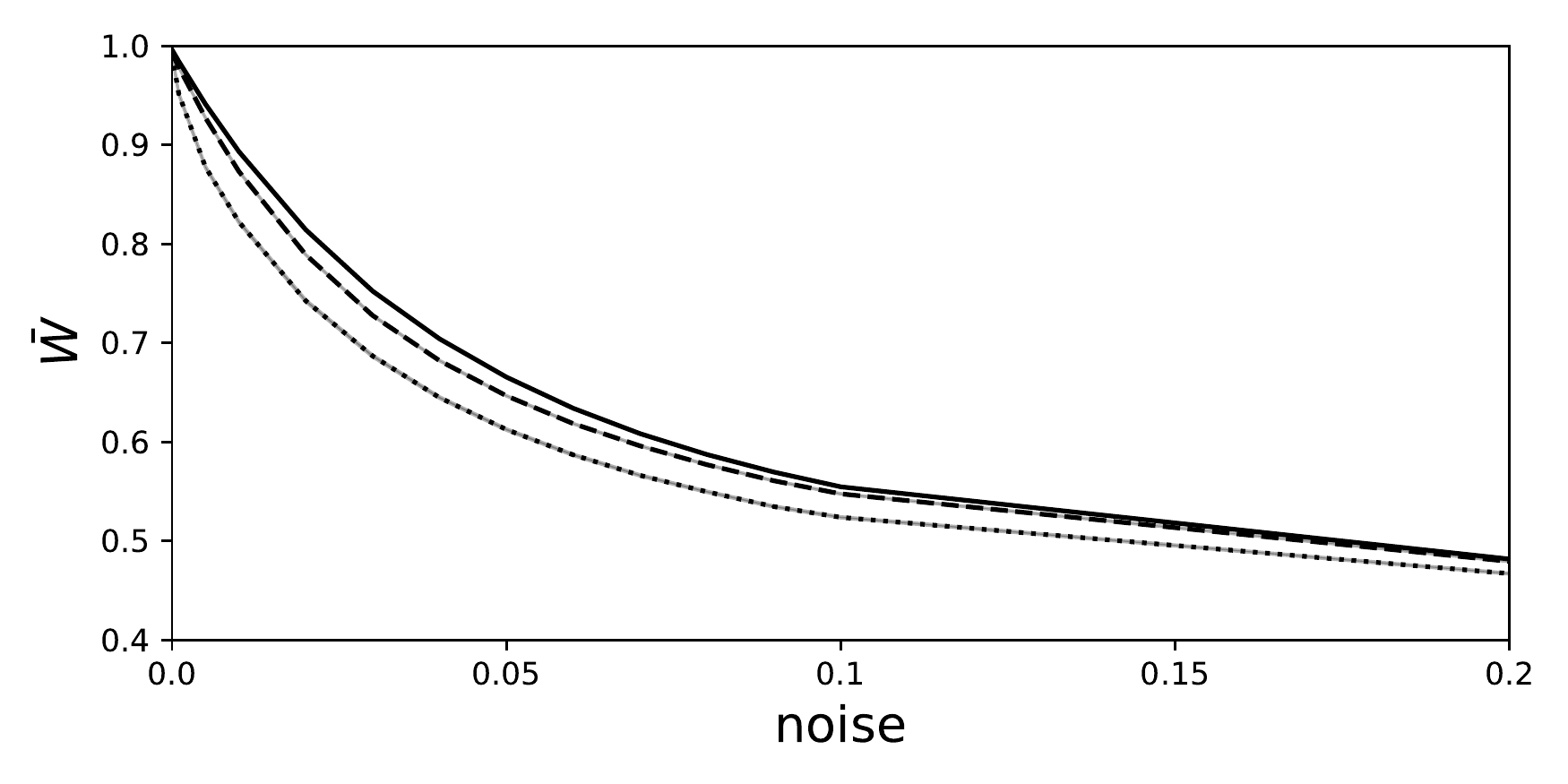} 
\end{center}
\caption{Robustness to sensor noise for the three different brain types. Robustness (mean performance) of Markov Brains (solid line), LSTMs (dashed line), and RNNs (dotted line) over noise. Error bars indicate the standard error.
}
\label{fig:robustness}
\end{figure}

Now that we have shown that indeed representations are more smeared in LSTMs and RNNs than they are in Markov Brains, we will ask how this smeardness relates to performance and robustness. After all, it has been speculated that the deep-learned convolutional networks are easily fooled~\citep{Szegedyetal2014} because the processing of information is spread out over all nodes. 
%After all, it has been speculated that the distributed nature of representations is the reason why deep-learned convolutional networks are easily fooled~\citep{Szegedyetal2014}. 
%Not sure that Szegedy talk about representations. I think that was my hypothesis [CA] but here it sounds we are attributing this to Szegedy.
We therefore compare how robust the different brain types are against sensor noise. We find that as expected, Markov Brains are the most robust, while LSTMs and RNNs are increasingly less robust against this type of noise (see Figure \ref{fig:robustness}).
%Fig. 12

In addition we find that robustness negatively correlates with smearedness across the different brain types (see Figure \ref{fig:smeardnessVsRobustness}). This is another indicator that more smeared representation do indeed make brains more vulnerable to noise, and that systems like Markov Brains, which have more discrete representations, also have an additional advantage due to the robustness of their representations.
%Fig. 13

\section{Discussion.}
A hallmark of intelligence is to reconstruct reality even from a very small set of information. One obvious way to attain this is to use sensory data to trigger models of the world that are stored within a brain, to fill in the missing parts. The information-theoretic concept of representations  quantifies precisely this capacity.

Creating machines that perform reality reconstruction using minimal sensory data is a daunting task, to the point where it has even been suggested that we would be better off creating machines without representations~\citep{Brooks_91}. 
%The vulnerability of deep-learned convolutional networks contributes to this problem. 
Here we showed that Markov Brains, LSTMs, and RNNs can be evolved to perform tasks that require representations, however these substrates differ greatly in the way they store this information. LSTMs and RNNs smear concepts over their hidden states, as we hypothesized, and consequently they become vulnerable to externally applied noise. In a manner of speaking, those smeared representations are ``dense" within the neuronal space. In contrast, we suggest that it is precisely the {\em sparseness} of representations that insulates them from noise. We note that another substrate that utilizes sparse representation is the HTM model ~\citep{HawkinsBlakeslee2004}, and it would be interesting to test that substrate for its robustness under noise. 

Might the  vulnerability to fluctuations in ``surface statistics" also be linked to ``catastrophic forgetting"~\citep{mccloskey1989catastrophic,ratcliff1990connectionist}, another plague of conventional systems based on the ANN paradigm?
While we have not tested this aspect of information encoding, reasonable arguments can be made that dense representations might be responsible for catastrophic forgetting as well. This may happen when new concepts overlap those concepts that are already present in the network.

We should point out that we did not test deep-learned LSTMs or RNNs,  but instead used neuroevolution to optimize them, so a direct comparison between the representations that we measured in LSTMS and RNNS and those that might be expected in deep convolutionary networks is still outstanding. There are two reasons for using neuroevolution in this context. First, we wanted to compare these systems to Markov Brains, and as of yet, we do not have a deep learning technique for them even though MBs have the ability to learn autonomously~\citep{sheneman2017evolving}. Deep Q-learning might be a viable alternative, since it is able to train probabilistic models with few hidden states~\citep{Mnihetal2015}. Regardless, the second reason for us to not use deep learning (or other gradient descent related methods) is that it is not clear to us how one would even use this technique to solve the task studied here. Perhaps it is time that we recognize that while deep-learned convolutional ANNs are great classifiers with an exceptional range of applications, they may be the wrong systems to create embodied thinking machines that exist and act in time. Instead, we should focus our attention on neuro-evolutionary systems and methods.

\section{Acknowledgements.}
This work was supported in part by Michigan State University through computational resources provided by the Institute for Cyber-Enabled Research. This material is based in part upon work supported by the National Science Foundation under Cooperative Agreement No. DBI-0939454.

\footnotesize
\bibliographystyle{apalike}
\bibliography{AL18} % replace by the name of your .bib file

\begin{thebibliography}{}

\bibitem[Albantakis et~al., 2014]{Albantakisetal2014}
Albantakis, L., Hintze, A., Koch, C., Adami, C., and Tononi, G. (2014).
\newblock Evolution of integrated causal structures in animats exposed to
  environments of increasing complexity.
\newblock {\em PLoS Comput Biol}, 10(12):e1003966.

\bibitem[Beer, 1996]{Beer1996}
Beer, R. (1996).
\newblock Toward the evolution of dynamical neural networks for minimally
  cognitive behavior.
\newblock In {P. Maes et al.}, editor, {\em Proc. 4th Intern. Conf. on
  Simulation of Adaptive Behavior}, pages 421--429, Cambridge, MA. MIT Press.

\bibitem[Beer, 2003]{Beer2003}
Beer, R. (2003).
\newblock The dynamics of active categorical perception in an evolved model
  agent.
\newblock {\em Adaptive Behavior}, 11:209--243.

\bibitem[Bengio et~al., 2015]{Bengioetal2015}
Bengio, Y., LeCun, Y., and Hinton, G. (2015).
\newblock Deep learning.
\newblock {\em Nature}, 521:436--444.

\bibitem[Bohm et~al., 2017]{bohm2017mabe}
Bohm, C., CG, N., and Hintze, A. (2017).
\newblock {MABE} (modular agent based evolver): {A} framework for digital
  evolution research.
\newblock In {C. Knibbe et al.}, editor, {\em Proceedings of the 14th European
  Conference on Artificial Life}, pages 76--83.

\bibitem[Brooks, 1991]{Brooks_91}
Brooks, R.~A. (1991).
\newblock Intelligence without representation.
\newblock {\em Artificial Intelligence}, 47:139--159.

\bibitem[Clark, 1997]{Clark1997a}
Clark, A. (1997).
\newblock The dynamical challenge.
\newblock {\em Cognitive Science}, 21:461--481.

\bibitem[George and Hawkins, 2009]{GeorgeHawkins2009}
George, D. and Hawkins, J. (2009).
\newblock Towards a mathematical theory of cortical micro-circuits.
\newblock {\em PLoS Comput Biol}, 5(10):e1000532.

\bibitem[Goodfellow et~al., 2016]{Goodfellowetal2016}
Goodfellow, I., Bengio, Y., and Courville, A. (2016).
\newblock {\em Deep Learning}.
\newblock MIT Press, Cambridge, MA.

\bibitem[Haugeland, 1991]{Haugeland1991}
Haugeland, J. (1991).
\newblock Representational genera.
\newblock In Ramsey, W., Stich, S.~P., and Rumelhart, D.~E., editors, {\em
  Philosophy and connectionist theory}, pages 61--89, Hillsdale, NJ. L.
  Erlbaum.

\bibitem[Hawkins and Blakeslee, 2004]{HawkinsBlakeslee2004}
Hawkins, J. and Blakeslee, S. (2004).
\newblock {\em On Intelligence}.
\newblock Henry Holt and Co., New York, NY.

\bibitem[Hintze et~al., 2017]{hintze2017markov}
Hintze, A., Edlund, J.~A., Olson, R.~S., Knoester, D.~B., Schossau, J.,
  Albantakis, L., Tehrani-Saleh, A., Kvam, P., Sheneman, L., Goldsby, H.,
  et~al. (2017).
\newblock Markov brains: {A} technical introduction.
\newblock {\em arXiv preprint arXiv:1709.05601}.

\bibitem[Hochreiter and Schmidhuber, 1997]{HochreiterSchmidhuber1997}
Hochreiter, S. and Schmidhuber, J. (1997).
\newblock Long short-term memory.
\newblock {\em Neural Computation}, 9(8):1735--1780.

\bibitem[Jo and Bengio, 2018]{JoBengio2018}
Jo, J. and Bengio, Y. (2018).
\newblock Measuring the tendency of {CNNs} to learn surface stastistical
  regularities.
\newblock arXiv:1711.11561.

\bibitem[Johnson-Laird and Wason, 1977]{johnson-laird}
Johnson-Laird, P. and Wason, P. (1977).
\newblock {\em Thinking: Readings in Cognitive Science}.
\newblock Cambridge University Press.

\bibitem[Lenski et~al., 2003]{lenski2003evolutionary}
Lenski, R.~E., Ofria, C., Pennock, R.~T., and Adami, C. (2003).
\newblock The evolutionary origin of complex features.
\newblock {\em Nature}, 423:139--144.

\bibitem[Marstaller et~al., 2013]{marstaller2013evolution}
Marstaller, L., Hintze, A., and Adami, C. (2013).
\newblock The evolution of representation in simple cognitive networks.
\newblock {\em Neural computation}, 25:2079--2107.

\bibitem[McCloskey and Cohen, 1989]{mccloskey1989catastrophic}
McCloskey, M. and Cohen, N.~J. (1989).
\newblock Catastrophic interference in connectionist networks: The sequential
  learning problem.
\newblock In {\em Psychology of Learning and Motivation}, volume~24, pages
  109--165. Elsevier.

\bibitem[Mnih et~al., 2015]{Mnihetal2015}
Mnih, V., Kavukcuoglu, K., Silver, D., Rusu, A.~A., and {J. Veness et al.}
  (2015).
\newblock Human-level control through deep reinforcement learning.
\newblock {\em Nature}, 518(7540):529--33.

\bibitem[Nguyen et~al., 2015]{Nguyenetal2015}
Nguyen, A., Yosinski, J., and Clune, J. (2015).
\newblock Deep neural networks are easily fooled: High confidence predictions
  for unrecognizable images.
\newblock In {\em Computer Vision and Pattern Recognition (CVPR '15)}. IEEE.

\bibitem[Phillips and Singer, 1997]{PhillipsSinger1997}
Phillips, W. and Singer, W. (1997).
\newblock In search of common foundations for cortical computation.
\newblock {\em Behav Brain Sci}, 20:657--683.

\bibitem[Phillips et~al., 1994]{Phillipsetal1994}
Phillips, W.~A., Kay, J., and Smyth, D.~M. (1994).
\newblock How local cortical processors that maximize coherent variation could
  lay foundations for representation proper.
\newblock In Smith, L.~S. and Hancock, P. J.~B., editors, {\em Neural
  Computation and Psychology}, pages 117--136, New York. Springer Verlag.

\bibitem[Pinker, 1989]{pinker}
Pinker, S. (1989).
\newblock {\em Learnability and Cognition}.
\newblock Cambridge, Mass.: The MIT Press.

\bibitem[Ratcliff, 1990]{ratcliff1990connectionist}
Ratcliff, R. (1990).
\newblock Connectionist models of recognition memory: constraints imposed by
  learning and forgetting functions.
\newblock {\em Psych Rev}, 97:285.

\bibitem[Russell et~al., 2003]{russell2003artificial}
Russell, S.~J., Norvig, P., Canny, J.~F., Malik, J.~M., and Edwards, D.~D.
  (2003).
\newblock {\em Artificial intelligence: {A} modern approach}.
\newblock Prentice Hall, Upper Saddle River.

\bibitem[Schmidhuber, 2015]{schmidhuber2015deep}
Schmidhuber, J. (2015).
\newblock Deep learning in neural networks: {A}n overview.
\newblock {\em Neural networks}, 61:85--117.

\bibitem[Schossau et~al., 2015]{schossau2015information}
Schossau, J., Adami, C., and Hintze, A. (2015).
\newblock Information-theoretic neuro-correlates boost evolution of cognitive
  systems.
\newblock {\em Entropy}, 18(1):6.

\bibitem[Sheneman and Hintze, 2017]{sheneman2017evolving}
Sheneman, L. and Hintze, A. (2017).
\newblock Evolving autonomous learning in cognitive networks.
\newblock {\em Scientific reports}, 7(1):16712.

\bibitem[Szegedy et~al., 2014]{Szegedyetal2014}
Szegedy, C., Zaremba, W., Sutskever, I., Bruna, J., Erhan, D., Goodfellow, I.,
  and Fergus, R. (2014).
\newblock Intriguing properties of neural networks.
\newblock In {\em International Conference on Learning Representations}.

\bibitem[{van Dartel}, 2005]{vandarteldiss}
{van Dartel}, M. (2005).
\newblock {\em Situated Representation}.
\newblock PhD thesis, Maastricht University.

\bibitem[{van Dartel} et~al., 2005]{vandarteletal05}
{van Dartel}, M., Sprinkhuizen-Kuyper, I., Postma, E., and {van den Herik}, J.
  (2005).
\newblock Reactive agents and perceptual ambiguity.
\newblock {\em Adaptive Behavior}, 13:227--42.

\end{thebibliography}

\end{document}